%% file: iclr2021_conference.tex
\documentclass{article} 
\usepackage{iclr2021_conference,times}

\input{math_commands.tex}

\usepackage{hyperref}
\usepackage{url}

\title{Class-conditioned Domain Generalization via Wasserstein Distributional Robust Optimization }

\author{Jingge Wang
\thanks{Jingge Wang, Yang Li (corresponding). 
Tsinghua-Berkeley Shenzhen Institute, Tsinghua University.
Emails: \texttt{wangjg19@mails.tsinghua.edu.cn, yangli@sz.tsinghua.edu.cn}
}, ~Yang Li\footnotemark[1], ~Liyan Xie\thanks{Liyan Xie, Yao Xie. School of Industrial and Systems Engineering, Georgia Institute of Technology. Emails:  \texttt{lxie49@gatech.edu, yao.xie@isye.gatech.edu} }, ~Yao Xie\footnotemark[2]
}  

\iclrfinalcopy 
\begin{document}

\maketitle

\begin{abstract}

Given multiple source domains, domain generalization aims at learning a universal model that performs well on any unseen but related target domain. In this work, we focus on the domain generalization scenario where domain shifts occur among class-conditional distributions of different domains. Existing approaches are not sufficiently robust when the variation of conditional distributions given the same class is large. In this work, we extend the 
concept of distributional robust optimization to solve the class-conditional domain generalization problem. Our approach optimizes the worst-case performance of a classifier over  class-conditional distributions within a Wasserstein ball centered around the barycenter of the source conditional distributions. We also propose an iterative algorithm for learning the optimal radius of the Wasserstein balls automatically. Experiments show that the proposed framework has better performance on unseen target domain than approaches without domain generalization.

\end{abstract}

\section{Introduction}
The distribution shift between training and testing data, a.k.a. domain shift, is a common problem in many realistic applications. One way to alleviate the adverse impact of domain shift is through domain generalization, which aims to learn 
a universal model based on available source datasets and in total absence of target data \citep{ghifary2016scatter, motiian2017unified}. 
Most existing methods learn a domain-invariant representation on source domains \citep{muandet2013domain,ghifary2015domain, motiian2017unified, li2018domain, li2018domain2}. However, these methods may encounter problems in certain  challenging domain generalization scenarios. 
Consider a data model $Y \rightarrow X$ defined on $\mathcal{X}\times\mathcal{Y}$. Given class label $Y$, feature $X$ is generated by conditional distributions $D_y(X)=P_D(X|Y=y)$ where $D$ denotes either a source domain ($S_m, m=1,\dots,M$) or the target domain $T$.
For a fixed $y\in \mathcal{Y}$, most domain generalization methods assume the class-conditional of target domain $T(X|Y=y)$ is closer to at least one of the $M$ source class-conditionals $S_m(X|Y=y)$ than to any distributions of another class (Figure \ref{intro} Left). In an ideal case, even simple $k$-NN based method can perform well. However, when the variation among class-conditionals of the same class is large, i.e., the closest conditional distribution to $T(X|Y=y)$ is some $S(X|Y=y^{\prime})$ of class $y^{\prime}\neq y$ (Figure \ref{intro} Right), aforementioned methods may not perform well \citep{krueger2020out}. 

In this work, we propose a class-conditioned domain generalization method 
inspired by the concept of distributional robust optimization \citep{kuhn2019wasserstein}, which optimizes the worst-case performance of a hypothesis over a set of distributions, namely the uncertainty set centered around the observed reference distributions. In the domain generalization context, we assume that class conditionals of different domains form class-specific uncertainty sets, aiming to learn a universally robust classifier that is even discriminative over the worst-case distributions in these sets. One challenge is to define a robust and computable uncertainty set.
When there is only one source domain, \citet{gao2018robust} defined the uncertainty set using Wasserstein distance and formulated the robust optimization problem as a convex optimization problem which can be solved efficiently. However, in domain generalization, there is no clear candidate for the reference distribution of each uncertainty set. Moreover, the radius of the Wasserstein uncertainty set, a fixed hyper-parameter in \citet{gao2018robust}, can largely impact the generalization performance. Our method uses Wasserstein barycenter as the reference distribution and an iterative algorithm for learning the optimal radius automatically.

\section{Background}
\citet{gao2018robust} formulates the robust hypothesis testing as a minimax problem considering distributionally uncertainty, i.e. given two sets of distributions over $X$, $\mathcal{P}_1$ and $\mathcal{P}_2$, find distributions $P_1\in \mathcal{P}_1,P_2\in\mathcal{P}_2$ and detector $\phi$ that minimizes the maximum of type I and type II error by solving
\begin{equation}
    \min _{\phi:\mathcal{X}\rightarrow \mathbb{R}}\max \bigg\{\sup _{P_{1} \in \mathcal{P}_{1}} P_{1}\{x: \phi(x)<0\}, \sup _{P_{2} \in \mathcal{P}_{2}} P_{2}\{x: \phi(x)\ge 0\}\bigg\}.\label{1}
\end{equation}

While there are various choices for the uncertainty sets, this work uses the Wasserstein distance to construct uncertainty ball of radius $\theta_k$ centered around empirical distribution $Q_k$, i.e., 
\( 
\mathcal{P}_{k}=\left\{P: \mathcal{W}\left(P, Q_k\right) \leq \theta_k\right\}, k = 1, 2.\label{2}
\) 
Problem (\ref{1}) can be transformed into an equivalent convex optimization problem of the following form  
\begin{equation}
\min _{\phi} \max _{P_{1} \in \mathcal{P}_{1}, P_{2} \in \mathcal{P}_{2}} \Phi\left(\phi; P_{1}, P_{2}\right),\label{3}
\end{equation}
where $\Phi$ represents the risk under certain distribution $P_1$ and $P_2$. After interchanging the $\min$ and $\max$ operators, we can first get optimal detector $\phi^*$ for any given $(P_1, P_2)$.
Then what is left is optimizing the least-favorable distributions (LFDs) $P_1^*$ and $P_2^*$ with uncertainty set constraints. Reformulating Wasserstein metric constraints in (\ref{2}) into equivalent linear constraints, the problem is finally transformed into a convex optimization problem. 

\section{Class-conditioned Domain Generalization}
Suppose we have access to $M$ diverse source domain with labeled data $\{(X_s^{m},Y_s^m)\}, m=1,\cdots,M$ which are representative of an underlying universal domain. 
Let $S_m(X|Y=y), m=1,\cdots,M$ and $T(X|Y=y)$ denote class-conditional distributions for each class $y$ in source and target domain, respectively.  
\begin{figure}
\CenterFloatBoxes
\begin{floatrow}
\ffigbox[0.6\textwidth]
  {\includegraphics[height = 2.8cm]{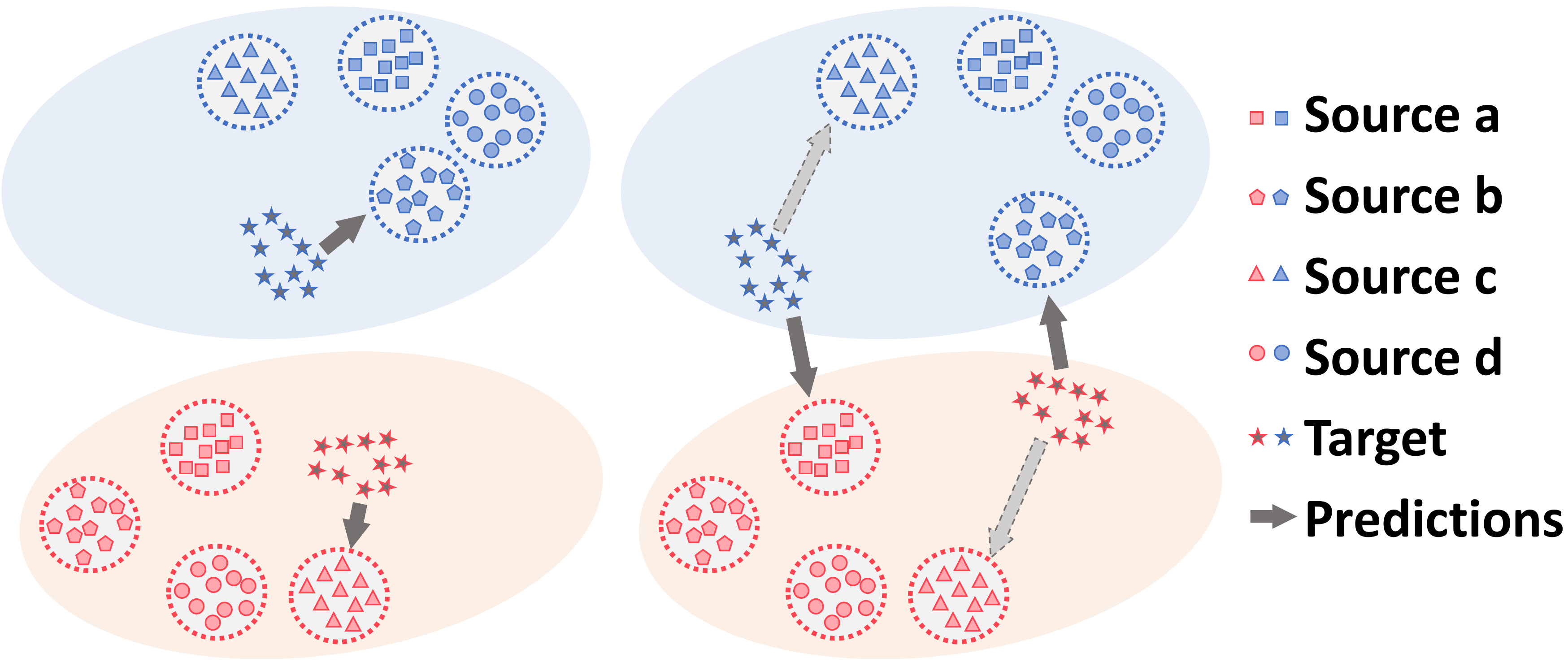}}
  {\caption{Classic and special setting with respect to class-conditional distributions (shaded circles indicate tasks, red and blue indicates different categories, best viewed in color. }\label{intro}}
\killfloatstyle
\ffigbox[0.38\textwidth]
  {\includegraphics[height = 2.8cm]{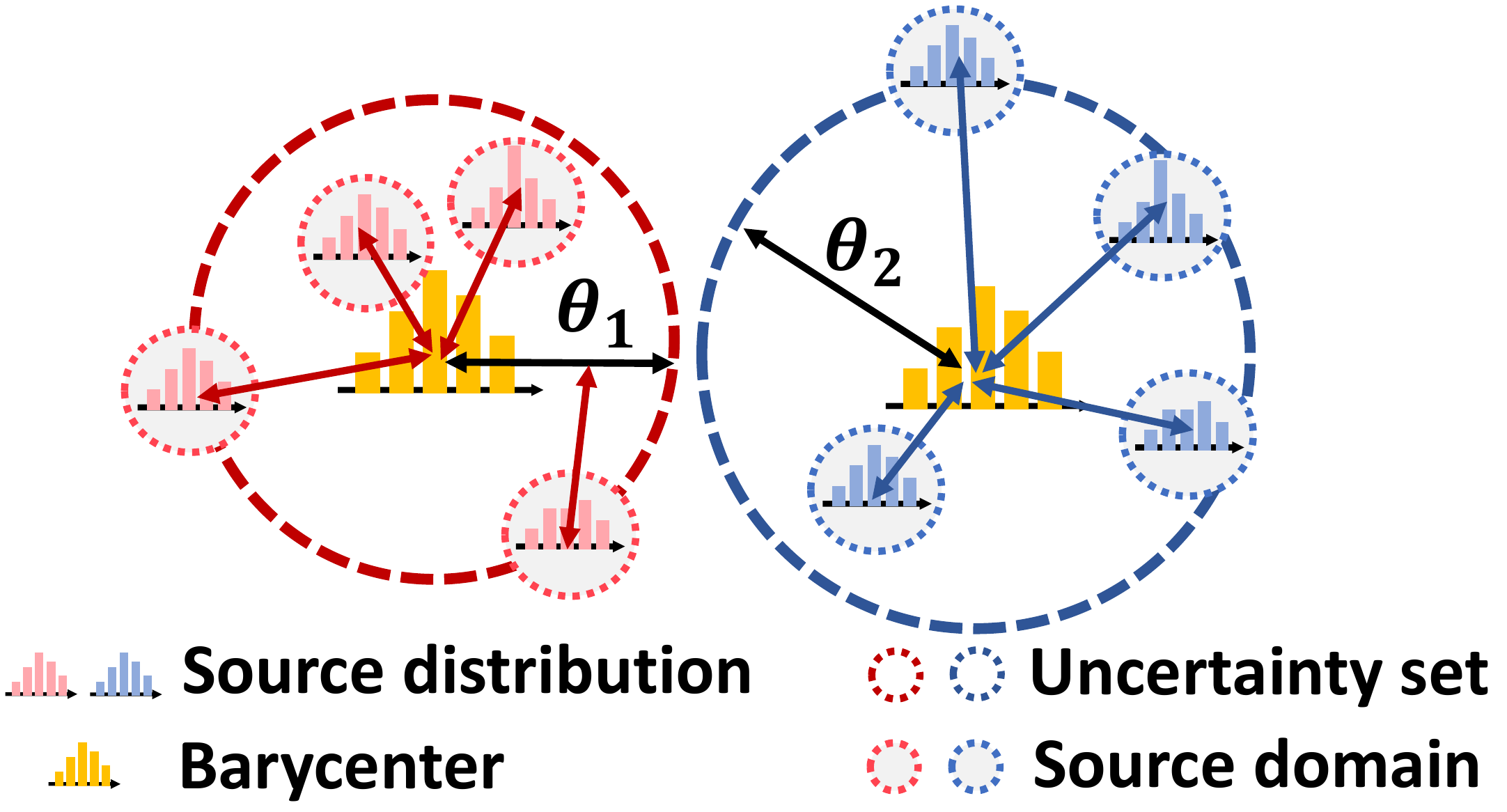}}
  {\caption{Initialization for $\theta_1, \theta_2$ using Wasserstein barycenter of source domain distributions.}\label{method}}
\end{floatrow}
\end{figure}
To construct class-specific uncertainty sets, the reference distribution and the radius need to be decided. 
Our method uses Wasserstein barycenter as the reference distribution, and introduces an iterative algorithm for learning the optimal radius.
Without loss of generality, we consider the binary classification setting. 

\textbf{Estimation of Distribution Uncertainty Sets.}
Without access to any target domain data, we can no longer use the empirical distribution as the center of uncertainty sets. A natural choice for the center of an uncertainty set defined by the source domains is the 2-Wasserstein barycenter since it better capture the geometry among distributions \citep{rabin2011wasserstein}.
Max value of all $M$ distances between source distributions and the barycenter is taken as the initial uncertainty set radius. 
Therefore, we can construct the uncertainty Wasserstein ball $\mathcal{P}_{1}, \mathcal{P}_{2}$ of radius $\theta_1$, $\theta_2$, which can be seen as general class-conditioned domain, as shown in Figure \ref{method}.

\textbf{Inference on Target Domain.}
Setting the generating function as exponential, the corresponding optimal detector is $\phi^*={1}/{2}\log({P_1}/{P_2})$ \citep{gao2018robust}.
Using the uncertainty sets of radius $\theta_1$, $\theta_2$, problem (\ref{3}) can produce worst-case distributions $P^*_1, P^*_2$, which are non-parametric functions of barycenter samples. To make inference on any target sample $x_t$, we define a weighted $k$-NN classifier as follows
\begin{equation}
\phi^{*}(x_t)= 
\begin{cases} 
1, & \text {if $\frac{1}{K}\sum_{i=1}^K w_i\log\left({P^*_1(x_i)}/{P^*_2(x_i)} \right)\ge0 $} \\ 
2, & \text{if $\frac{1}{K}\sum_{i=1}^K w_i\log\left({P^*_1(x_i)}/{P^*_2(x_i)} \right)<0 $},
\end{cases}
\end{equation}
where $x_1,\dots,x_K$ are the K nearest neighbors of $x_t$ measured by the  Euclidean distance and $w_i$ is inversely proportional to $\left\|x_i - x_t\right\|_2$. We use $K=3$ in the experiments. 

\textbf{Dynamically Learning of $\theta_1, \theta_2$.}
In practice, the initial radius of Wasserstein balls tends to be too large and
the optimal class-conditionals $P_1^*$ and $P_2^*$ may become indistinguishable.
Thus we add one more constraint to ensure the LFDs are significantly different, using the  chi-squared test, whose p-value is denoted as $\rho_\epsilon\left(P_1(\theta_1), P_2(\theta_2)\right)$, 
and the problem becomes as follows
\begin{equation}
\begin{aligned}
    & \min _{\phi} \max _{P_{1} \in \mathcal{P}_{1}(\theta_1), P_{2} \in \mathcal{P}_{2}(\theta_2)} \Phi\left(\phi; P_{1}, P_{2}\right) \\
    & \text{s.t.} \ \rho_\epsilon\left(P_1(\theta_1), P_2(\theta_2)\right) < \gamma,\\
\end{aligned}
\end{equation}
where $\gamma$ is the significance threshold taken as $0.05$ in Chi-square testing.
To avoid the difficulty of solving the optimization problem with non-convex constraint directly, we use a heuristic method to search for the optimal radius satisfying the constraint. 
As larger uncertainty set leads to less distinguishable LFDs, we initialize $\theta_1, \theta_2$ as the maximum Wasserstein distance between source distributions and the barycenter distribution. In each iteration, $\theta_1,\theta_2$ are dynamically decremented, i.e., $\theta_1 = \theta_1 - \Delta, \theta_2 = \theta_2 - \Delta$ with a small positive $\Delta$, until ${P}_1$ and ${P}_2$ are significantly different.
More details can be found in Algorithm \ref{algorithm}.

\renewcommand{\algorithmicrequire}{\textbf{Input:}}  
\renewcommand{\algorithmicensure}{\textbf{Output:}} 

\begin{algorithm}[t]
\caption {Learning algorithm for class-conditioned domain generalization} \label{algorithm}
\begin{algorithmic}[0]
\Require $M$ diverse source tasks with labeled data $\mathcal{X}_{s}^m=\{(X_s^{m},Y_s^m)\}, m = 1,\cdots,M$;
\Ensure The LFDs $P_1^* ,P_2^*$ supported on source task samples;
\State {1. Initialization of $\theta_1,\theta_2$:}
\For{each class $y$}
    \State Barycenter distribution $C^{*}(X|y) \gets \argmin _{C(X|y)} \sum_{m=1}^{M} \frac{1}{M} \mathcal{W}_{2}^{2}\left(C(X|y), S_m(X|y)\right)$;
    \State Initial uncertainty set radius $\theta_y \gets \max\limits_{m=1,\cdots,M}\mathcal{W}_{2}\left(C(X|y), S_m(X|y)\right)$;
\EndFor{}
\State {2. Dynamically Learning of $\theta_1,\theta_2$:}
\Repeat
\State la
\State $\theta_1 \gets \theta_1 - \Delta, \theta_2 \gets \theta_2 - \Delta$;
\Until $ \rho_\epsilon(P_1^*, P_2^*) < \gamma$
\end{algorithmic}
\end{algorithm}

\section{Experimental Evaluation}
\begin{figure}
\CenterFloatBoxes
\begin{floatrow}
\ffigbox[0.4\textwidth]
  {\includegraphics[height = 2.3cm]{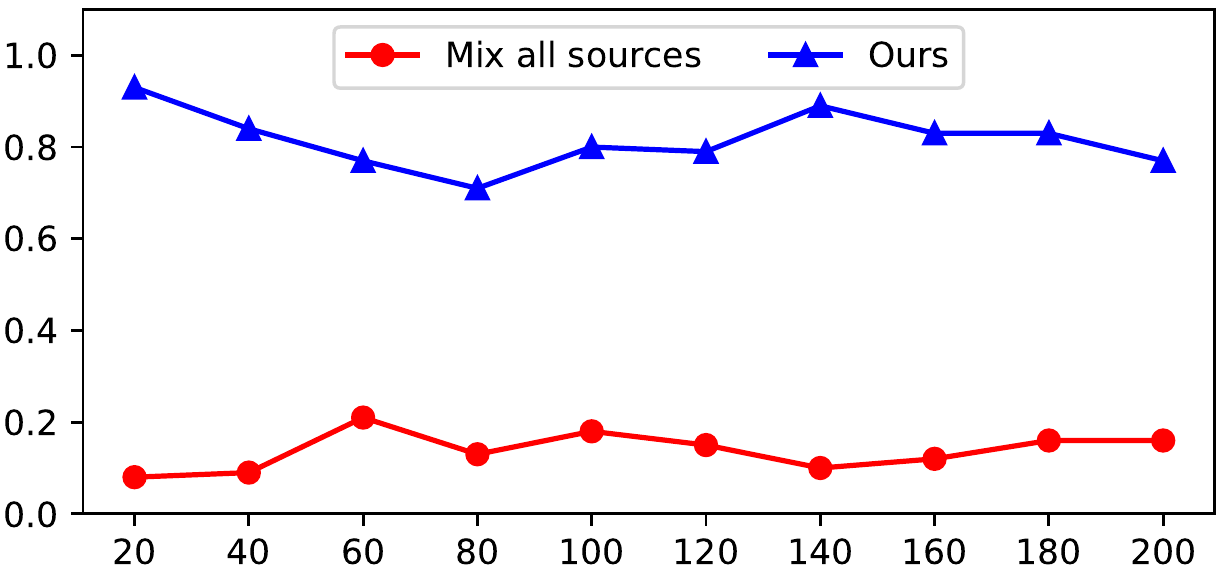}}
  {\caption{Average accuracy of unseen Gaussian data from target domain using source Gaussian data with different sample size. }\label{gaussian}}
\killfloatstyle
\ttabbox[0.55\textwidth]
{\small
  \setlength{\tabcolsep}{0.3mm}
{\scalebox{0.8}{
  \begin{tabular}{ccccccc}
\toprule
\multirow{2}{*}{} & \multicolumn{6}{c}{Difficultiy level} \\
\cmidrule{2-7}
&{1}&{2}&{3}&{4}&{5}&{6}\\
\midrule
\multirow{2}{*}{Capacity range} 
&[0.95,1] &[0.945,1] &[0.94,1] &[0.93,1] &[0.92,1] &[0.905,1] \\
&[0,0.85] &[0,0.855] &[0,0.86] &[0,0.87] &[0,0.88] &[0,0.895] \\\midrule
Training &16 &18 &20 &24 &28 &38 \\
Testing  &84 &92 &102 &122 &146 &196  \\\midrule
$\theta_1$ &3.44 &3.33 &3.36 &3.20 &3.04 &2.65\\
$\theta_2$ &4.02 &3.96 &3.63 &3.04 &2.80 &2.37\\
\bottomrule
\end{tabular}}}
  }
{\caption{The range of for each class, data size and initial uncertainty set radius in varying difficulty settings for battery datasets.}
\label{data-size}}
\end{floatrow}
\end{figure}

\textbf{Synthesized Data.} We first evaluate our algorithm on data generated from Gaussian-like class-conditional distributions for each source and target domain. Setting four source domains and one target domain configured in a way similar to Figure 1 (right), we vary the source sample size from 20 to 200 and fix the target testing sample size at 60. 
Results shown in Figure \ref{gaussian} indicate that regardless of different sample size settings, the $k$-NN classifier trained by mixing all the source domain data shows far worse performance on the target testing data compared with our method. This simple experiment illustrates the potential of handling special data scenario of our method.

\floatsetup[table]{capposition=top}
\newfloatcommand{capbtabbox}{table}[][\FBwidth]
\begin{table}
\small\setlength{\tabcolsep}{1.5mm}
\scalebox{0.9}
{
\begin{tabular}{cccccccc}
\toprule
\multirow{2}{*}{} 
&\multirow{2}{*}{Method} 
& \multicolumn{6}{c}{Difficulty level} \\
\cmidrule{3-8}
& & {1}&{2}&{3}&{4}&{5}&{6}\\
\midrule
\multirow{5}{*}{Single domain } 
& Target only (supervised) &\bf0.935 &0.927 & 0.925 &\bf0.899 &0.850 &\bf0.782 \\
& Source a only (unsupervised)    & {0.929} &\bf0.934 &\bf0.927  &0.884 &\bf0.872 &0.761  \\
&  Source b only (unsupervised)    &0.224 &0.218 &0.184 &0.184 &0.228 &0.270  \\
& Source c only (unsupervised)    &0.918 &0.927 &0.909 & 0.866 &0.822 &0.751 \\
&  Source d only (unsupervised)    &0.083 &0.085 &0.078 & 0.108 &0.160 &0.247 \\ \midrule
\multirow{4}{*}{Domain generalization} 
& Source a+b+c+d (unsupervised)   &0.525 &0.538 &0.451 &0.476  &0.469 &0.478  \\
& Source a+b+c+d+target (semi-supervised) &0.680 &0.691 &0.626 &0.609 &0.601 & {\bf 0.568}  \\
& Ours w/o radius learning(unsupervised) &0.740 &0.670 &0.608 &0.402 &0.599 &0.517\\
& Ours w/ radius learning(unsupervised) &\textbf{0.806} &\textbf{0.795} & \textbf{0.765} &\textbf{0.681}  &\textbf{0.628} & 0.530  \\
\bottomrule
\end{tabular}
}
\caption{Comparison of binary classification accuracy under 6 difficulty grade settings.}
\label{results}
\end{table}

\textbf{Real-world Data.}
We adopt five datasets collected from electric charge-discharge tests of power batteries under different experimental conditions. Using one dataset as the target domain and the other four as source domains, our goal is to decide whether a battery should be retired based on its charge-discharge features. 
For each sample, 17-dimension features represents the lab testing result, and binary label represents the battery state determined by its capacity level.
By varying the capacity range of each battery state, we create six experimental scenarios with different classification difficulty levels denoted by numbers 1-6 in Table \ref{data-size}. The higher the level, the harder the task. 
To simulate the realistic setting of scarce labeled data, in each trial we randomly sample 1/10 of the training set as source domain data and calculate the average accuracy over 20 trials for all experiments. 
Sampled training and testing sizes for each source domain are also shown in Table \ref{data-size}. 

We compare our method and its truncated version without dynamically learning of $\theta$ on target testing data with the following baselines based on $k$-NN: (1) Target only: use $k$-NN on target training data; (2) Source a/b/c/d only: use data from one of the four source domains; (3) Source a+b+c+d: mix all source training data; (4) Source a+b+c+d+target: mix source and target training data together. The truncated version uses the initial $\theta$ shown in the last two rows in Table \ref{data-size}, which is used as initialization in the standard version.

Results in Table \ref{results} shows that our proposed method outperforms the approach of mixing available data in both unsupervised and semi-supervised way. It is inferior to two source only methods but outperforms the other two. 
This shows the usability of our method especially when we have no idea which source is more similar with the target in an unsupervised setting. 
The truncated version only yields competitive results 0.740 compared with semi-supervised approach in the easiest grade, implying the necessity of learning for $\theta$. 

\section{Conclusions}
In this paper, we present a robust domain generalization method that can effectively learn a universal classifier invariant to domain shifts in the class conditional distributions. Testing on both synthesized and real-world data, we show that this method has promising performance using only limited source domains.
In the future, we will extend our method to more learning scenarios, such as multi-class classification, and evaluate our framework on more real-world applications.


  \subsubsection*{Acknowledgments}
   This research is supported by Natural Science Foundation of China 62001266 and partially supported by NSF CAREER CCF-1650913. We thank Aihua Ran, Shuxiao Chen, Zihao Zhou and Guodan Wei from Tsinghua-Berkeley Shenzhen Institute for providing the battery data.

\bibliography{iclr2021_conference}
\bibliographystyle{iclr2021_conference}


\end{document}

%% file: math_commands.tex
\usepackage{amsmath,amsfonts,bm}

\def\eqref#1{equation~\ref{#1}}

\def\1{\bm{1}}

\DeclareMathAlphabet{\mathsfit}{\encodingdefault}{\sfdefault}{m}{sl}
\SetMathAlphabet{\mathsfit}{bold}{\encodingdefault}{\sfdefault}{bx}{n}

\DeclareMathOperator*{\argmin}{arg\,min}